\documentclass{article}
\usepackage{authblk}
\usepackage{times}
\usepackage{helvet}
\usepackage{courier}
\usepackage{url}
\usepackage{graphicx}
\usepackage{stmaryrd}

\usepackage{amssymb}
\usepackage{xspace}
\usepackage{color}
\usepackage{subcaption}
\usepackage{graphicx}
\usepackage{amssymb}
\usepackage{amsmath,amsthm,mathtools}
\usepackage{tabu}
\usepackage{booktabs}
\usepackage{adjustbox}
\usepackage{array}
\newcommand{\dashedline}{\tabucline[0.3pt on 3pt]\\}
\usepackage{mathabx}
\usepackage{multirow}

\theoremstyle{definition}
\newtheorem{definition}{Definition}
\theoremstyle{remark}

\usepackage{colonequals}

\newcommand{\tuple}[1]{\ensuremath{\langle #1 \rangle}}
\usepackage{bm}

\newcommand{\slop}[1]{\ensuremath{\omega_{#1}}}
\newcommand{\slbel}[1]{\ensuremath{b_{#1}}}
\newcommand{\sldis}[1]{\ensuremath{d_{#1}}}
\newcommand{\slunc}[1]{\ensuremath{u_{#1}}}
\newcommand{\slbase}[1]{\ensuremath{a_{#1}}}
\newcommand{\sloptuple}[1]{\ensuremath{\tuple{\slbel{#1}, \sldis{#1}, \slunc{#1}, \slbase{#1}}}}
\newcommand{\variance}[1]{\ensuremath{\sigma^2_{#1}}}
\newcommand{\mean}[1]{\ensuremath{\mu_{#1}}}
\newcommand{\var}[1]{\variance{#1}}

\newcommand{\semiring}{\ensuremath{\mathfrak{S}}}
\newcommand{\semiringp}{\ensuremath{\semiring_p}}
\newcommand{\semiringsl}{\ensuremath{\semiring_{\mbox{SL}}}}
\newcommand{\semiringbeta}{\ensuremath{\semiring^{\beta}}}

\title{Probabilistic Logic Programming with Beta-Distributed Random Variables\thanks{This research was sponsored by the U.S. Army Research Laboratory and the U.K. Ministry of Defence under Agreement Number W911NF-16-3-0001. The views and conclusions contained in this document are those of the authors and should not be interpreted as representing the official policies, either expressed or implied, of the U.S. Army Research Laboratory, the U.S. Government, the U.K. Ministry of Defence or the U.K. Government. The U.S. and U.K. Governments are authorized to reproduce and distribute reprints for Government purposes notwithstanding any copyright notation hereon.}}
\author{Federico Cerutti\\Cardiff University\\ \and Lance Kaplan\\ ARL\\ \and Angelika Kimmig \\ Cardiff University\\ \and Murat \c{S}ensoy\\ Ozyegin University}
\begin{document}
\maketitle

\begin{abstract}
We enable aProbLog---a probabilistic logical programming approach---to reason in presence of uncertain probabilities represented as Beta-distributed random variables. We achieve the same performance of state-of-the-art algorithms for highly specified and engineered domains, while simultaneously we maintain the flexibility offered by aProbLog in handling complex relational domains.
Our motivation is that faithfully capturing the distribution of probabilities is necessary to compute an expected utility for effective decision making under uncertainty: unfortunately, these probability distributions can be highly uncertain due to sparse data. To understand and accurately manipulate such probability distributions we need a well-defined theoretical framework that is provided by the Beta distribution, which specifies a distribution of probabilities representing all the possible values of a probability when the exact value is unknown.
\end{abstract}

\section{Introduction}
In the last years, several probabilistic variants of Prolog
have been developed, such as 
ICL~\cite{Poole00}, Dyna~\cite{Eisner05}, PRISM~\cite{Sato01} and ProbLog~\cite{deraedt:ijcai07}, with its aProbLog extension \cite{DBLP:conf/aaai/KimmigBR11} to handle arbitrary labels from a semiring (Section \ref{sec:aproblog}).
They all are based on
definite clause logic (pure Prolog) extended with facts labelled with  probability values.
Their meaning is typically derived from
Sato's distribution semantics~\cite{sato:iclp95}, which assigns a probability to every literal. The probability of a Herbrand interpretation, or possible world, is the product of the probabilities of the literals occurring in this world.
The success probability is the probability that a query succeeds in a randomly selected world.

Faithfully capturing the distribution of the probabilities of such queries is necessary for effective decision making under uncertainty to compute an expected utility \cite{von2007theory}. Often such distributions are learned from prior experiences that can be provided either by  subject matter experts or by objective recordings.

Unfortunately, these probability distributions can be highly uncertain and this significantly affects decision making \cite{anderson2016using,antonucci2014decision}. In fact, not all scenarios are blessed with a substantial amount of data enabling reasonable characterisation of probability distributions. For instance, when dealing with adversarial behaviours such as policing operations, training data is sparse or subject matter experts have limited experience to elicit the probabilities. 

To understand and accurately manipulate such probability distributions, we need a well-defined theoretical framework that is provided by the Beta distribution, which specifies a distribution of probabilities representing all the possible values of a probability when the exact value is unknown. This has been recently investigated in the context of singly-connected Bayesian Network, in an approach named Subjective Bayesian Network (SBN) \cite{ivanovska.15,kaplan.16.fusion,KAPLAN2018132}, that shows higher performance against other traditional approaches dealing with uncertain probabilities, such as Dempster-Shafer Theory of Evidence \cite{DEMPSTER68,Smets2005}, and  replacing single probability values with closed intervals representing the possible range of probability values \cite{credal98}. SBN is based on Subjective Logic \cite{Josang2016-SL} (Section \ref{sec:betabackground}) that provides an alternative, more intuitive, representation of Beta distributions as well as a calculus for manipulating them. Subjective logic has been successfully applied in a variety of domains, from trust and reputation \cite{josang2006trust}, to urban water management \cite{MOGLIA2012180}, to assessing the confidence of  neural networks for image classification \cite{Sensoy2018}.

In this paper, we enable aProbLog \cite{DBLP:conf/aaai/KimmigBR11} to reason in presence of uncertain probabilities represented as Beta distribution. Among other features, aProbLog is freely available\footnote{\url{https://dtai.cs.kuleuven.be/problog/}} and it directly handles  Bayesian networks,\footnote{As pointed out by \cite{fierens:tplp15}, for such Bayesian network models, ProbLog inference is tightly linked to the inference approach of \cite{Sang2005}.} which simplifies our experimental setting when comparing against SBN and other approaches on Bayesian Networks with uncertain probabilities.
We determine a parametrisation for aProbLog (Section \ref{sec:beta-operators}) deriving operators for addition, multiplication, and division operating on Beta-distributed random variables matching the results to a new Beta-distributed random variable using the moment matching method \cite{Minka2001,KLEITER1996143,VANALLEN2008483,KAPLAN2018132}.

We achieve the same results of highly engineered approaches for inferencing in single-connected Bayesian networks---in particular in presence of high uncertainty in the distribution of probabilities which is our main research focus---and simultaneously we maintain the flexibility offered by aProbLog in handling complex relational domains. 
Results of our experimental analysis (Section \ref{sec:experiment}) indeed indicate that the proposed approach (1) handles inferences in general aProbLog programs better than using standard subjective logic operators \cite{Josang2016-SL} (Appendix \ref{sec:sl-operators}), and (2) it performs equivalently to state-of-the-art approaches of reasoning with uncertain probabilities \cite{KAPLAN2018132,credal98,Smets2005}, despite the fact that they have been highly engineered for the specific case of single connected Bayesian Networks while we can handle  general aProbLog programs.

\section{Background}

\subsection{aProbLog}
\label{sec:aproblog}

For a set $J$ of ground facts, we define the set of literals $\operatorname{L}(J)$ and the set of interpretations $\mathcal{I}(J)$ as follows:
\begin{align}
\operatorname{L}(J) &= J\cup\{\neg f~|~f\in J \}\\
\mathcal{I}(J) &= \{S~|~S\subseteq \operatorname{L}(J) \wedge \forall l\in J:~l\in S\leftrightarrow \neg l \notin S\}
\end{align}

An algebraic Prolog (aProbLog) program \cite{DBLP:conf/aaai/KimmigBR11} consists of:
\begin{itemize}
\item a \emph{commutative semiring} $\tuple{\mathcal{A},\oplus,\otimes, e^{\oplus},e^{\otimes}}$\footnote{That is, \emph{addition}~$\oplus$ and \emph{multiplication}~$\otimes$ are associative and commutative binary operations over the set $\mathcal{A}$, $\otimes$ distributes over $\oplus$, $e^{\oplus}\in\mathcal{A}$ is the neutral element with respect to~$\oplus$, $e^{\otimes}\in\mathcal{A}$ that of~$\otimes$, and for all $a\in \mathcal{A}$, $e^{\oplus}\otimes a = a \otimes e^{\oplus} = e^{\oplus}$.}
\item a finite set of ground \emph{algebraic facts} $\operatorname{F} = \{ f_1, \ldots, f_n\}$
\item a finite set $\operatorname{BK}$ of \emph{background knowledge clauses}
\item a \emph{labeling function} $\delta : \operatorname{L}(\operatorname{F})\rightarrow \mathcal{A}$
\end{itemize}
Background knowledge clauses are definite clauses, but their bodies may contain negative literals for algebraic facts. Their heads may not unify with any algebraic fact.

For instance, in the following aProbLog program
\begin{verbatim}
alarm :- burglary.
0.05 :: burglary.
\end{verbatim}

\noindent
\verb+burglary+ is an algebraic fact with label \verb+0.05+, and \verb+alarm :- burglary+ represents a background knowledge clause, whose intuitive meaning is: in case of burglary, the alarm should go off.

The idea of splitting a logic program in a set of facts and a set of clauses goes back to Sato's distribution semantics~\cite{sato:iclp95}, where it is used to define a probability distribution over interpretations of the entire program in terms of a distribution over the facts.
This is possible because a truth value assignment to the facts in $\operatorname{F}$ uniquely determines the truth values of all other atoms defined in the background knowledge.
In the simplest case, as realised in ProbLog \cite{deraedt:ijcai07,fierens:tplp15}, this basic distribution considers facts to be independent random variables and thus multiplies their individual probabilities. aProbLog uses the same basic idea, but generalises from the semiring of probabilities to general commutative semirings.
While the distribution semantics is defined for countably infinite sets of facts,
the set of ground algebraic facts in aProbLog must be finite.

In aProbLog, the label of
a complete interpretation $I\in \mathcal{I}(\operatorname{F})$  is defined as the product of the labels of its literals
\begin{equation}
\operatorname{\mathbf{A}}(I)  =  \bigotimes_{l\in I}\delta(l) \label{eq:w_expl}
\end{equation}
and the label of a set of interpretations $S\subseteq \mathcal{I}(\operatorname{F})$ as the sum of the interpretation labels
\begin{equation}
\operatorname{\mathbf{A}}(S)  = \bigoplus_{I\in S}\bigotimes_{l\in I}\delta(l)
\end{equation}
A \emph{query} $q$ is a finite set of algebraic literals and atoms from the Herbrand base,\footnote{I.e., the set of ground atoms that can be constructed from the predicate, functor and constant symbols of the program.} $q\subseteq \operatorname{L}(\operatorname{F}) \cup HB(\operatorname{F}\cup \operatorname{BK})$. We denote the set of interpretations where the query is true by $\mathcal{I}(q)$,
\begin{equation}
\mathcal{I}(q) = \{I~|~I\in \mathcal{I}(\operatorname{F}) \wedge I\cup \operatorname{BK}\models q\}
\end{equation}
The label of query $q$ is defined as the label of $\mathcal{I}(q)$,
\begin{equation}
\operatorname{\mathbf{A}}(q)  =  \operatorname{\mathbf{A}}(\mathcal{I}(q)) = \bigoplus_{I\in \mathcal{I}(q)}\bigotimes_{l\in I}\delta(l).\label{eq:q_int}
\end{equation}
As both operators are commutative and associative, the label is independent of the order of both literals and interpretations.

In the context of this paper, we extend aProbLog to queries with evidence by introducing an additional division operator~$\oslash$ that defines the conditional label of a query as follows:
\begin{equation}
\label{eq:fusion}
\displaystyle{\mathbf{A}(q|\bm{E}=\bm{e})} = \mathbf{A}(\mathcal{I}(q ~\land~ \bm{E}=\bm{e})) ~\oslash~\mathbf{A}(\mathcal{I}(\bm{E}=\bm{e}))
\end{equation}

\noindent
where $\mathbf{A}(\mathcal{I}(q ~\land~ \bm{E}=\bm{e})) ~\oslash~ \mathbf{A}(\mathcal{I}(\bm{E}=\bm{e}))$ returns the label of $q ~\land~ \bm{E}=\bm{e}$ given the label of $\bm{E} = \bm{e}$. We refer to a specific choice of semiring, labeling function and division operator as an \emph{aProbLog parametrisation}.

ProbLog is an instance of aProbLog with the following parameterisation, which we denote $\semiringp$: 
\begin{equation}
    \label{eq:semiringprobability}
    \begin{array}{l}
         \mathcal{A} = \mathbb{R}_{\geq 0};\\
         a ~\oplus~ b = a + b;\\
         a ~\otimes~ b = a \cdot b;\\
         e^\oplus = 0;\\
         e^{\otimes} = 1;\\
         \delta(f) \in [0,1];\\
         \delta(\lnot f) = 1 - \delta(f);\\
         a ~\oslash~ b = \frac{a}{b}
    \end{array}
\end{equation}

\noindent

\subsection{Beta Distribution and Subjective Logic Opinions}
\label{sec:betabackground}

When probabilities are uncertain---for instance because of limited observations---such an uncertainty can be captured by a Beta distribution, namely a distribution of possible probabilities. Let us consider only binary variables such as $X$ that can take on the value of true  or false, i.e., $X= {x}$ or $X =\bar{{x}}$.  The value of $X$ does change over different instantiations, and there is an underlying ground truth value for the probability $p_x$ that $X$ is true ($p_{\bar{x}} = 1 - p_x$  that $X$ is false). 
If $p_x$ is drawn from a Beta distribution, it has the following probability density function:
\begin{equation}
\label{eq:Dirichlet distribution}
f_{\bm{\beta}}(p_x;\bm{\alpha})= \frac{1}{\beta(\alpha_x,\alpha_{\bar{x}})}p_x^{\alpha_x-1}(1-p_x)^{\alpha_{\bar{x}}-1}\;
\end{equation}
for $0 \le p_x \le 1$, where $\beta(\cdot)$ is the beta function and the beta parameters are
$\bm{\alpha}_X = \tuple{\alpha_x,\alpha_{\bar{x}}}$, such that $\alpha_x > 1, \alpha_{\bar{x}} > 1$. Given a Beta-distributed random variable $X$, 
\begin{equation}
\label{eq:strength}
s_X = \alpha_x + \alpha_{\bar{x}}
\end{equation}
 is its \emph{Dirichlet strength} and 
\begin{equation}
    \label{eq:mean}
    \mu_X = \frac{\alpha_x}{s_X}
\end{equation}
is its mean. From \eqref{eq:strength} and \eqref{eq:mean} the beta parameters can  equivalently be written as:
\begin{equation}
    \label{eq:parameterssx}
    \bm{\alpha_X} = \tuple{\mu_X s_X, ~(1-\mu_X) s_X} .
\end{equation}
The variance of a Beta-distributed random variable $X$ is
\begin{equation}
\label{e:pred_var}
\sigma_X^2 = \frac{\mu_X (1 - \mu_X)}{s_X+1}
\end{equation}
and from \eqref{e:pred_var} we can rewrite $s_X$ \eqref{eq:strength} as
\begin{equation}
\label{eq:sxvar}
s_X = \frac{\mu_X (1-\mu_X)}{\sigma^2_X} - 1 .
\end{equation}

\subsubsection{Parameter Estimation}
Given a random variable $Z$ with known mean $\mean{Z}$ and variance $\var{Z}$, we can use the method of moments and \eqref{eq:sxvar} to estimate the $\bm{\alpha}$ parameters of a Beta-distributed variable $Z'$ of mean $\mean{Z'} = \mean{Z}$ and
\begin{equation}
    \label{eq:minvarcheck}
    s_{Z'} = \max\left\{\frac{\mu_Z (1-\mu_Z)}{\sigma^2_Z} -1, \frac{W \slbase{Z}}{\mu_Z}, \frac{W (1 - \slbase{Z})}{(1-\mu_Z)} \right\}.
\end{equation}
\eqref{eq:minvarcheck} is needed to ensure that the resulting Beta-distributed random variable $Z'$ does not lead to a $\bm{\alpha}_{Z'} \leq \tuple{1,1}$.

\subsubsection{Beta-Distributed Random Variables from Observations}
The value of $X$ can be observed from $N_{ins}$ independent observations of $X$. If over these observations, $n_x$ times $X = {x}$, $n_{\bar{x}} = N_{ins}-n_x$ times $X = \bar{{x}}$, then $\bm{\alpha}_{X} = \tuple{n_x + W \slbase{X}, n_{\bar{x}} + W (1 - \slbase{X})}$: $\slbase{X}$ is the prior assumption, i.e. the probability that $X$ is true in the absence of observations;
and $W > 0$ is a prior weight indicating the strength of the prior assumption. Unless specified otherwise, in the following we will assume $\forall X, \slbase{X} = 0.5$ and $W = 2$, so to have an uninformative, uniformly distributed, prior.

\subsubsection{Subjective Logic}
Subjective logic \cite{Josang2016-SL} provides (1) an alternative, more intuitive, way of representing the parameters of a Beta-distributed random variables, and (2) a set of operators for manipulating them. A subjective opinion about a proposition $X$ is a tuple $\slop{X} = \sloptuple{X}$, representing the belief,
disbelief and uncertainty that $X$ is true at a given instance, and, as above, $\slbase{X}$  is the prior probability that $X$ is true in the absence of observations. These values are non-negative and $\slbel{X} + \sldis{X} + \slunc{X} = 1$. The
projected probability $P(x) =\slbel{X} + \slunc{X} \cdot \slbase{X}$, provides an estimate of the ground truth probability $p_x$. 

The mapping from a Beta-distributed random variable $X$ with parameters $\bm{\alpha}_X = \tuple{\alpha_{x}, \alpha_{\bar{x}}}$ to a subjective opinion is:
\begin{equation}
    \slop{X} = \left\langle\frac{\alpha_x - W \slbase{X}}{s_X}, \frac{\alpha_{\bar{x}} - W (1 - \slbase{X})}{s_X}, \frac{W}{s_X}, \slbase{X}\right\rangle
\end{equation}
With this transformation, the mean of $X$ is equivalent to the projected probability $P(x)$, and the Dirichlet strength is inversely proportional to the uncertainty of the opinion:
\begin{equation}
    \mu_X = P(x) = \slbel{X} + \slunc{X} \slbase{X}, \quad s_X = \frac{W}{\slunc{X}}
\end{equation}

Conversely, a subjective opinion $\slop{X}$  translates directly into a Beta-distributed random variable with:
\begin{equation}
    \bm{\alpha}_X = \left\langle\frac{W}{\slunc{X}} \slbel{X} + W \slbase{X}, \frac{W}{\slunc{X}} \sldis{X} + W (1 - \slbase{X})\right\rangle
\end{equation}

Subjective logic is a framework that includes various operators to indirectly determine opinions from various logical operations. In particular, we will make use of $\boxplus_{SL}$, $\boxtimes_{SL}$, and $\boxslash_{SL}$, resp. summing, multiplying, and dividing two subjective opinions as they are defined in \cite{Josang2016-SL} (Appendix \ref{sec:sl-operators}).
Those operators aim at faithfully matching the projected probabilities: for instance the multiplication of two subjective opinions $\slop{X} \boxtimes_{SL} \slop{Y}$ results in an opinion $\slop{Z}$ such that $P(z) = P(x) \cdot P(z)$. 

The straightforward approach to derive a aProbLog parametrisation for operations in subjective logic is to use the operators $\boxplus$, $\boxtimes$, and $\boxslash$.

\begin{definition}
\label{def:semiringSL}
The aProbLog parametrisation $\semiringsl$ is defined as follows:
\begin{equation}
    \label{eq:semiringsloperators}
    \begin{array}{l}
          \mathcal{A_{\mbox{SL}}} = \mathbb{R}_{\geq 0}^4;\\
          a ~\oplus_{\mbox{SL}}~ b = a ~\boxplus_{\mbox{SL}}~ b;\\
          a ~\otimes_{\mbox{SL}}~ b = a ~\boxtimes_{\mbox{SL}} b;\\
          e^{\oplus_{\mbox{SL}}} =  \tuple{0, 1, 0, 0};\\
          e^{\otimes_{\mbox{SL}}} = \tuple{1, 0, 0, 1};\\
          \delta_{\mbox{SL}}(f_i) = \sloptuple{f_i} \in [0,1]^4;\\
          \delta_{\mbox{SL}}(\lnot f_i) = \tuple{\sldis{f_i}, \slbel{f_i}, \slunc{f_i}, 1 - \slbase{f_i}};\\
          a ~\oslash_{\mbox{SL}}~ b 
           = \left\{
          \begin{array}{l l}
            a ~\boxslash_{\mbox{SL}}~ b  & \mbox{if defined}\\
            \tuple{0,0,1,0.5} & \mbox{otherwise}
          \end{array}
          \right.
    \end{array}
\end{equation}
\end{definition}

Note that $\tuple{\mathcal{A_{\mbox{SL}}}, \oplus_{\mbox{SL}}, \otimes_{\mbox{SL}}, e^{\oplus_{\mbox{SL}}}, e^{\otimes_{\mbox{SL}}}}$ does not form a commutative semiring in general. If we consider only the projected probabilities---i.e. the means of the associated Beta distributions---then $\boxplus$ and $\boxtimes$ are indeed commutative, associative, and $\boxtimes$ distributes over $\boxplus$. However, the uncertainty of the resulting opinion depends on the order of operands.

\section{Operators for Beta-Distributed Random Variables}
\label{sec:beta-operators}

While SL operators try to faithfully characterise the projected probabilities, they employ an uncertainty maximisation principle to limit the belief commitments, hence they have a looser connection to the Beta distribution. The operators we derive in this section aim at maintaining such a connection.

Let us first define a sum operator between two independent Beta-distributed random variables $X$ and $Y$ as the Beta-distributed random variable $Z$ such that $\mean{Z} = \mean{X + Y}$ and $\variance{Z} = \variance{X + Y}$. The sum (and in the following the product as well) of two Beta random variables is not necessarily a Beta random variable. Our approach, consistent with \cite{KAPLAN2018132}, approximates the resulting distribution as a Beta distribution via moment matching on mean and variance: this guarantees to approximate the result as a Beta distribution. 

\begin{definition}[Sum]
\label{def:subbeta}
Given $X$ and $Y$ independent Beta-distributed random variables represented by the subjective opinion $\slop{X}$ and $\slop{Y}$, the \emph{sum} of $X$ and $Y$ ($\slop{X} \boxplus^\beta \slop{Y}$) is defined as the Beta-distributed random variable $Z$ such that:
    $\mean{Z} = \mean{X + Y} = \mean{X} + \mean{Y}$
and
$\variance{Z} = \variance{X + Y} = \variance{X} + \variance{Y}$.
\end{definition}

$\slop{Z} = \slop{X} \boxplus^\beta \slop{Y}$ can then be obtained as discussed in Section \ref{sec:betabackground}, taking \eqref{eq:minvarcheck} into consideration. The same applies for the following operators as well.

Let us now define the product operator between two independent Beta-distributed random variables $X$ and $Y$ as the Beta-distributed random variable $Z$ such that $\mean{Z} = \mean{XY}$ and $\variance{Z} = \variance{XY}$.

\begin{definition}[Product]
\label{def:productbeta}
Given $X$ and $Y$ independent Beta-distributed random variables represented by the subjective opinion $\slop{X}$ and $\slop{Y}$, the \emph{product} of $X$ and $Y$ ($\slop{X} \boxtimes^\beta \slop{Y}$) is defined as the Beta-distributed random variable $Z$ such that:
    $\mean{Z} = \mean{XY} = \mean{X} ~ \mean{Y}$
and 
        $\variance{Z} = \variance{XY} =  \variance{X} (\mean{Y})^2 + \variance{Y} (\mean{X})^2 + \variance{X} \variance{Y}$.
\end{definition}

Finally, let us define the conditioning-division operator between two independent Beta-distributed random variables $X$ and $Y$, represented by subjective opinions $\slop{X}$ and $\slop{Y}$, as the Beta-distributed random variable $Z$ such that $\mean{Z} = \mean{\frac{X}{Y}}$ and $\variance{Z} = \variance{\frac{X}{Y}}$. 

\begin{definition}[Conditioning-Division]
\label{def:product}
Given $\slop{X} = \sloptuple{X}$ and $\slop{Y} = \sloptuple{Y}$ subjective opinions such that X and Y are Beta-distributed random variables, $Y = \mathbf{A}(\mathcal{I}(\bm{E}=\bm{e})) = \mathbf{A}(\mathcal{I}(q ~\land~ \bm{E}=\bm{e})) ~\oplus~ \mathbf{A}(\mathcal{I}(\lnot q ~\land~ \bm{E}=\bm{e}))$, with $\mathbf{A}(\mathcal{I}(q ~\land~ \bm{E}=\bm{e})) = X$. The \emph{conditioning-division} of $X$ by $Y$ ($\slop{X} \boxslash^\beta \slop{Y}$) is defined as the Beta-distributed random variable $Z$ such that:\footnote{In the following, $\simeq$ highlights the fact that the results are obtained using the the first order Taylor approximation.}

\begin{equation}
    \begin{split}
        \mean{Z} = \mean{\frac{X}{Y}} = \mean{X} \mean{\frac{1}{Y}} \simeq \frac{\mean{X}}{\mean{Y}}
    \end{split}
\end{equation}
and 

\begin{equation}
    \label{eq:variance-division}
    \begin{split}
        \variance{Z} & \simeq  (\mean{Z})^2 (1 - \mean{Z})^2 \cdot \\
        & \cdot \left( \frac{\variance{X}}{(\mean{X})^2}
        + \frac{\variance{Y} - \variance{X}}{(\mean{Y} - \mean{X})^2} +  \frac{2 \variance{X}}{\mean{X} (\mean{Y} - \mean{X})}
        \right)
    \end{split}
\end{equation}
\end{definition}

We can now define a new aProbLog parametrisation similar to Definition \ref{def:semiringSL} operating with our newly defined operators $\boxplus^\beta$, $\boxtimes^\beta$, and $\boxslash^\beta$.

\begin{definition}
\label{def:semiringbeta}
The aProbLog parametrisation $\semiringbeta$ is defined as follows:
\begin{equation}
    \label{eq:semiringsloperatorsbeta}
    \begin{array}{l}
          \mathcal{A^\beta} = \mathbb{R}_{\geq 0}^4;\\
          a ~\oplus^\beta~ b = a ~\boxplus^\beta~ b;\\
          a ~\otimes^\beta~ b = a ~\boxtimes^\beta b;\\
          e^{\oplus^\beta} = \tuple{1, 0, 0, 0.5};\\
          e^{\otimes^\beta} = \tuple{0, 1, 0, 0.5};\\
          \delta^\beta(f_i) = \sloptuple{f_i} \in [0,1]^4;\\
          \delta^\beta(\lnot f_i) = \tuple{\sldis{f_i}, \slbel{f_i}, \slunc{f_i}, 1 - \slbase{f_i}};\\
          a ~\oslash^\beta~ b = a ~\boxslash^\beta~ b
    \end{array}
\end{equation}
\end{definition}

As per Definition \ref{def:semiringSL}, also $\tuple{\mathcal{A^\beta}, \oplus^{\beta}, \otimes^{\beta}, e^{\oplus^{\beta}}, e^{\otimes^{\beta}}}$ is not in general a commutative semiring. Means are correctly matched to projected probabilities, therefore for them $\semiringbeta$ actually operates as a semiring. However, for what concerns variance, the product is not distributive over addition:  $\var{X (Y + Z)} = \var{X} (\mean{Y} + \mean{Z})^2 + (\var{Y} + \var{Z}) \mean{X}^2 + \var{X} (\var{Y} + \var{Z}) \neq \var{X} (\mean{Y}^2 + \mean{Z}^2) + (\var{Y} + \var{Z}) \mean{X}^2 + \var{X} (\var{Y} + \var{Z}) = \var{(XY) + (XZ)}$. The approximation error we introduce is therefore
\begin{equation}
    \label{eq:approxerror}
    \begin{split}
    e(X, Y, Z) & \leq \\
    & \frac{2\mean{Y}\mean{Z} \var{X}}{\var{X} (\mean{Y}^2 + \mean{Z}^2) + (\mean{X}^2 + \var{X}) (\var{Y} + \var{Z})}
    \end{split}
\end{equation}
and it minimally affects the results both in the case of low and in the case of high uncertainty in the random variables.

\section{Experimental Analysis}
\label{sec:experiment}

\begin{figure*}[t]
\centering
    \begin{subfigure}[b]{0.3\textwidth}
        \includegraphics[width=\textwidth]{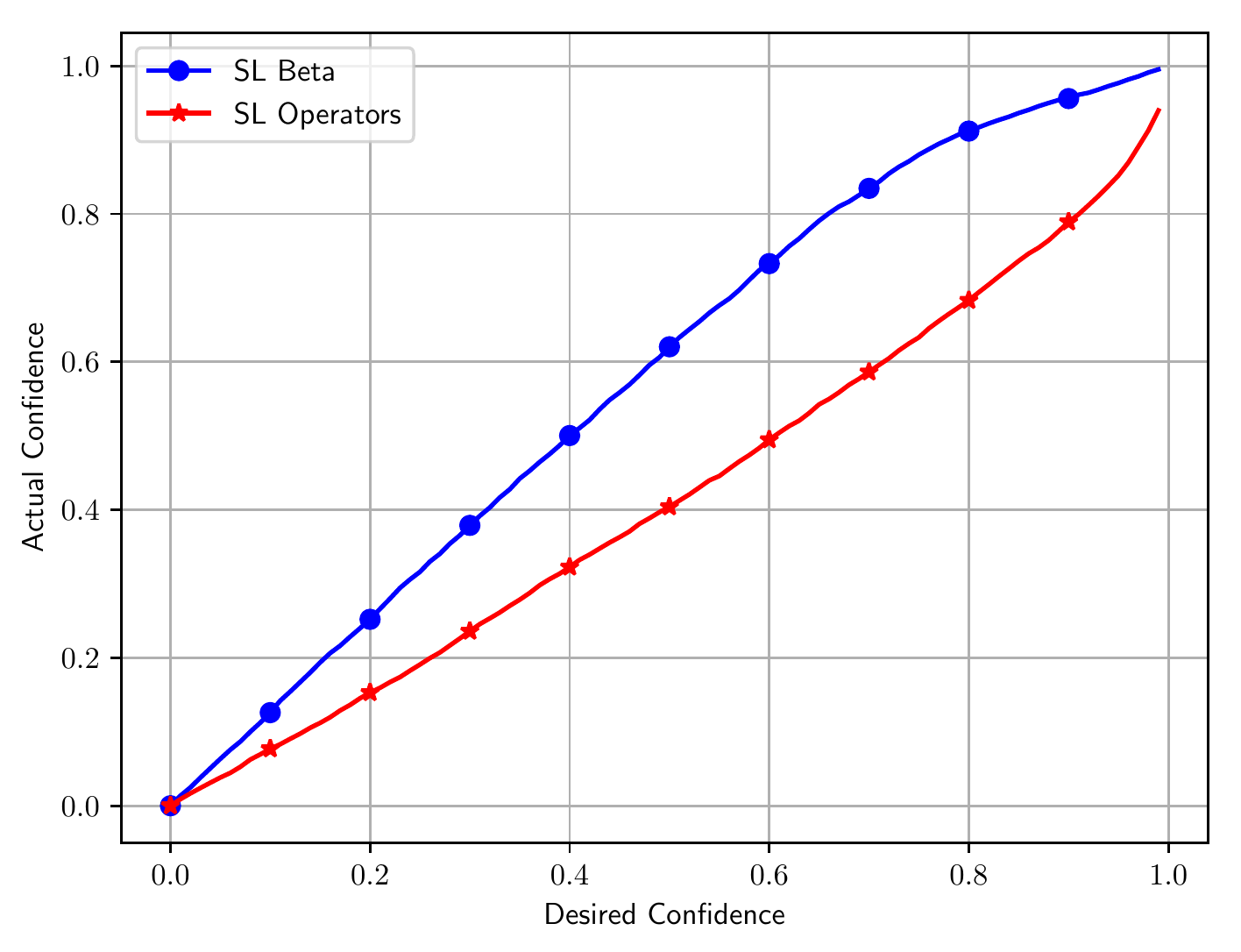}
        \caption{}
        \label{fig:sfa}
    \end{subfigure}
    ~
    \begin{subfigure}[b]{0.3\textwidth}
        \includegraphics[width=\textwidth]{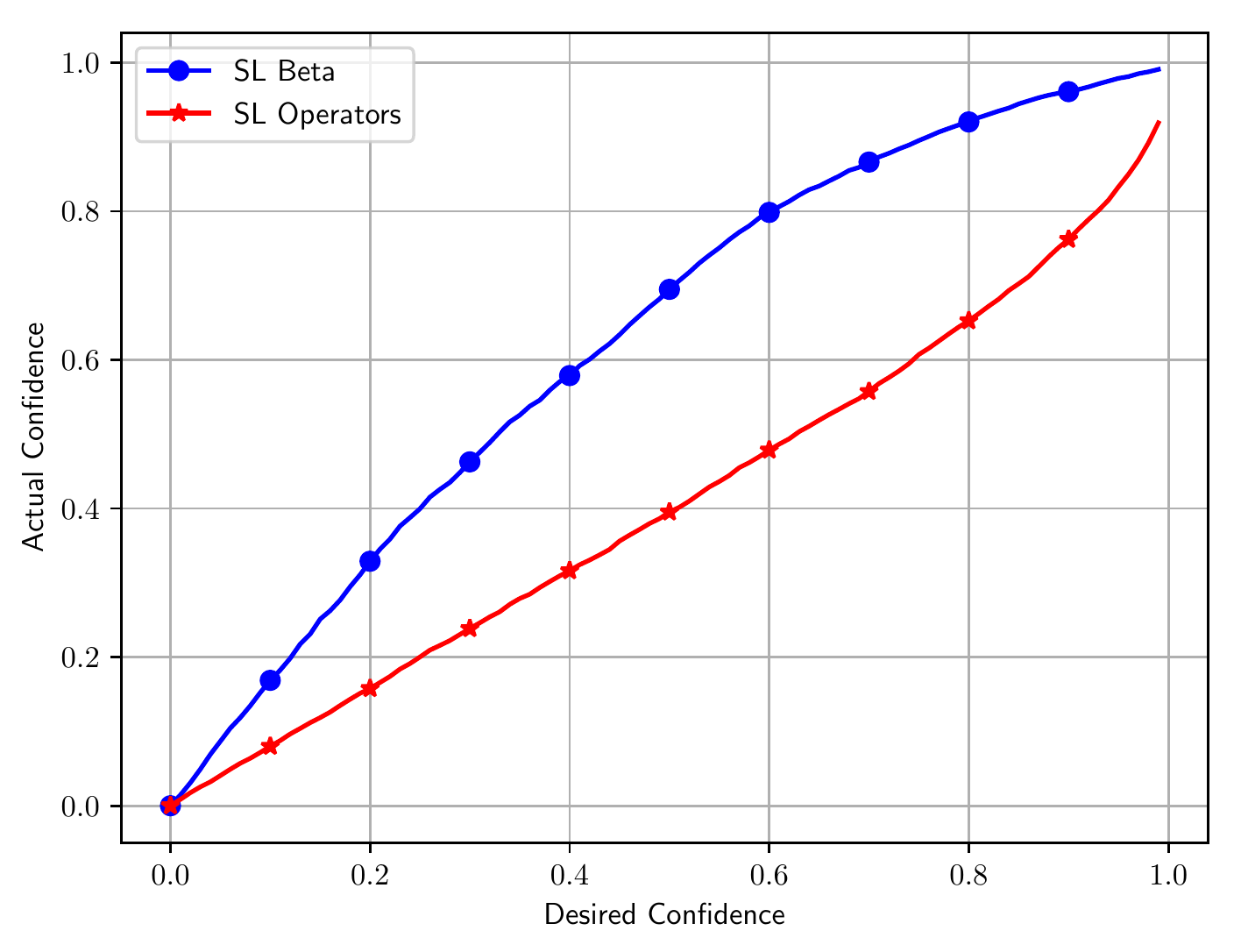}
        \caption{}
        \label{fig:sfb}
    \end{subfigure}
    ~
    \begin{subfigure}[b]{0.3\textwidth}
        \includegraphics[width=\textwidth]{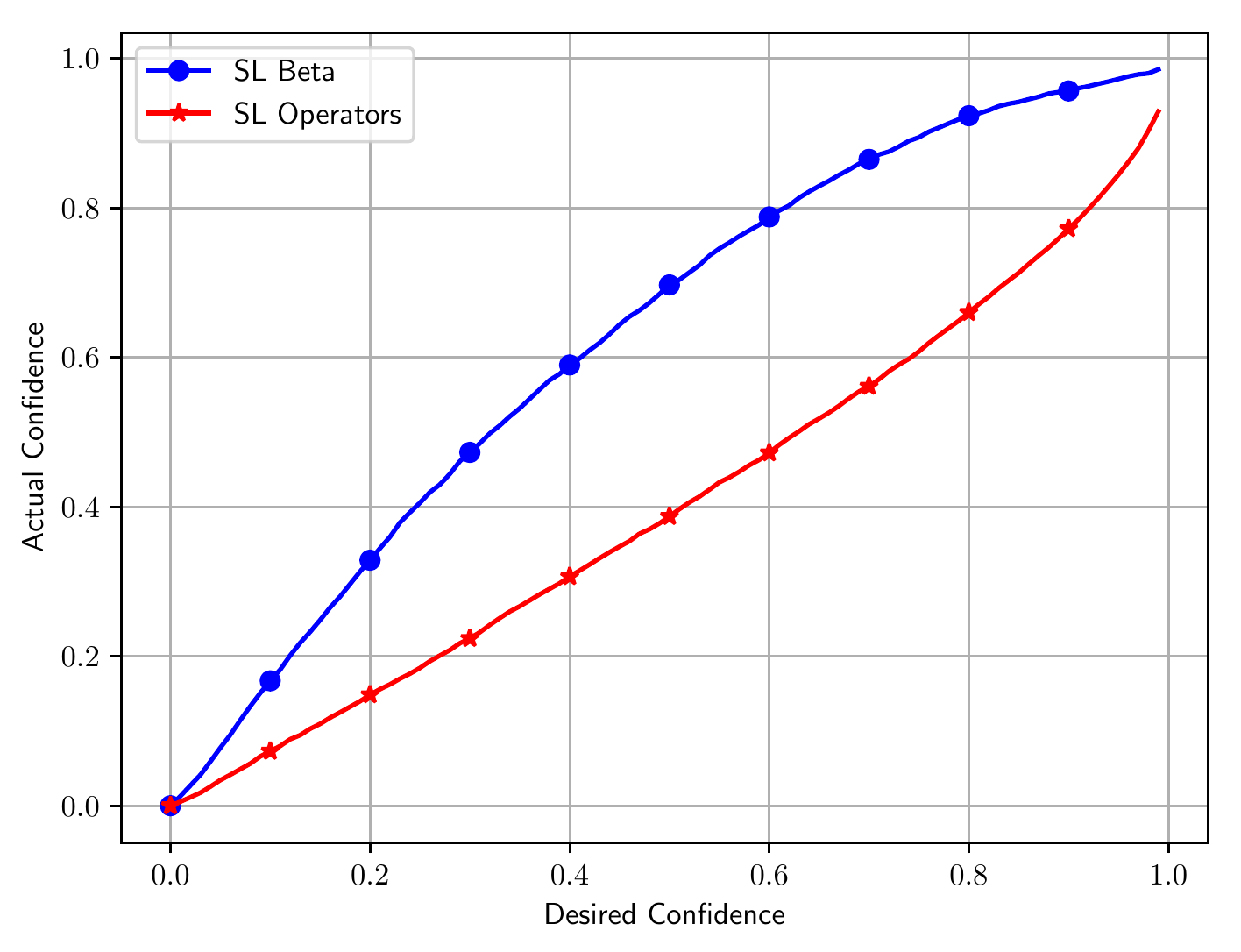}
        \caption{}
        \label{fig:sfc}
    \end{subfigure}
    
    \caption{Actual versus desired significance of bounds derived from the uncertainty for Smokers \& Friends with: (\subref{fig:sfa}) $N_{ins} = 10$; (\subref{fig:sfb} $N_{ins} = 50$; and (\subref{fig:sfc}) $N_{ins} = 100$. Best closest to the diagonal. In the figure, \emph{SL Beta} represents aProbLog with $\semiringbeta$, and \emph{SL Operators} represents aProbLog with $\semiringsl$. }
    \label{fig:sf}
\end{figure*}

To evaluate the suitability of using $\semiringbeta$ in aProbLog for uncertain probabilistic reasoning, we run an experimental analysis involving several aProbLog programs with unspecified labelling function. For each program, first labels are derived for $\semiringp$ by selecting the ground truth probabilities from a uniform random distribution. Then, for each label of the aProbLog program over $\semiringp$, we derive a subjective opinion by observing $N_{ins}$ instantiations of the random variables comprising the aProbLog program over $\semiringp$ so to simulate data sparsity \cite{KAPLAN2018132}.
We then proceed analysing the inference on specific query nodes $\bm{q}$ in the presence of a set of evidence $\bm{E}=\bm{e}$ using aProbLog with $\semiringsl$ and $\semiringbeta$ over the subjective opinion labels, and compare the RMSE to the actual ground truth of using aProbLog with $\semiringp$. This process of inference to determine the marginal Beta distributions is repeated 1000 times by considering 100 random choices for each label of the aProbLog with $\semiringp$, i.e. the ground truth,  and for each ground truth  10 repetitions of sampling the interpretations 
used to derive the subjective opinion labels used in $\semiringsl$ and $\semiringbeta$  observing $N_{ins}$ instantiations of all the variables. 

Following \cite{KAPLAN2018132}, we judge the quality of the Beta distributions of the queries on how well its expression of uncertainty captures the spread between its projected probability and the actual ground truth probability. In simulations where the ground truths are known, such as ours, confidence bounds  can be formed around the projected probabilities at a significance level of $\gamma$ and determine the fraction of cases when the ground truth falls within the bounds. If the uncertainty is well determined by the Beta distributions, then this fraction should correspond to the strength $\gamma$ of the confidence interval \cite[Appendix C]{KAPLAN2018132}. 

\subsection{Inferences in Arbitrary aProbLog Programs}

\begin{table}
    \small
    \centering
    \begin{tabu}{X[1,l] X[1,l] X[1,l] X[1,r] X[1,r]}
    \toprule
    Program        & $N_{ins}$ &  & $\semiringbeta$ & $\semiringsl$\\
    \midrule
    \multirow[t]{ 6}{=}{\setlength\parskip{\baselineskip}Friends \newline{} \& Smokers}           & 10  & Actual & \textbf{0.1014} & 0.1514 \\
    &                & Predicted    & 0.1727 & 0.1178 \\
               & 50  & Actual & \textbf{0.0620} & 0.1123 \\
                      & & Predicted & 0.0926 & 0.0815 \\
& 100 & Actual & \textbf{0.0641} & 0.1253 \\
& & Predicted & 0.1150 & 0.0893 \\
    \bottomrule
    
    \end{tabu}
    \caption{RMSE for the queried variables in the Friends \& Smokers program: best results for the actual RMSE in bold.}
    \label{tab:smokers}
\end{table}

We first considered the famous Friends \& Smokers problem\footnote{\url{https://dtai.cs.kuleuven.be/problog/tutorial/basic/05_smokers.html}} with fixed queries and set of evidence, to illustrate the behaviour between $\semiringsl$ and $\semiringbeta$. 
Table \ref{tab:smokers} provides the root mean square error (RMSE) between the projected probabilities and the ground truth probabilities for all the inferred query variables for $N_{ins}$ = 10, 50, 100. The table also includes the predicted RMSE by taking the square root of the average---over the number of runs---variances from the inferred marginal Beta distributions, cf. Eq. \eqref{e:pred_var}. Figure \ref{fig:sf} plots the desired and actual significance levels for the confidence intervals (best closest to the diagonal), i.e. the fractions of times the ground truth falls within confidence bounds set to capture x\% of the data. 

The aProbLog with $\semiringbeta$ exhibits the lowest RMSE, and is a little conservative in estimating its own RMSE, while aProbLog with $\semiringsl$ is overconfident. This reflects in Figure \ref{fig:sf}, with the results of aProbLog with $\semiringbeta$ being over the diagonal, and those of aProbLog with $\semiringsl$ being below it.

\subsection{Inferences in aProbLog Programs Representing Single-Connected Bayesian Networks}

\begin{figure*}[t]
\centering
    \begin{subfigure}[b]{0.25\textwidth}
        \includegraphics[width=\textwidth]{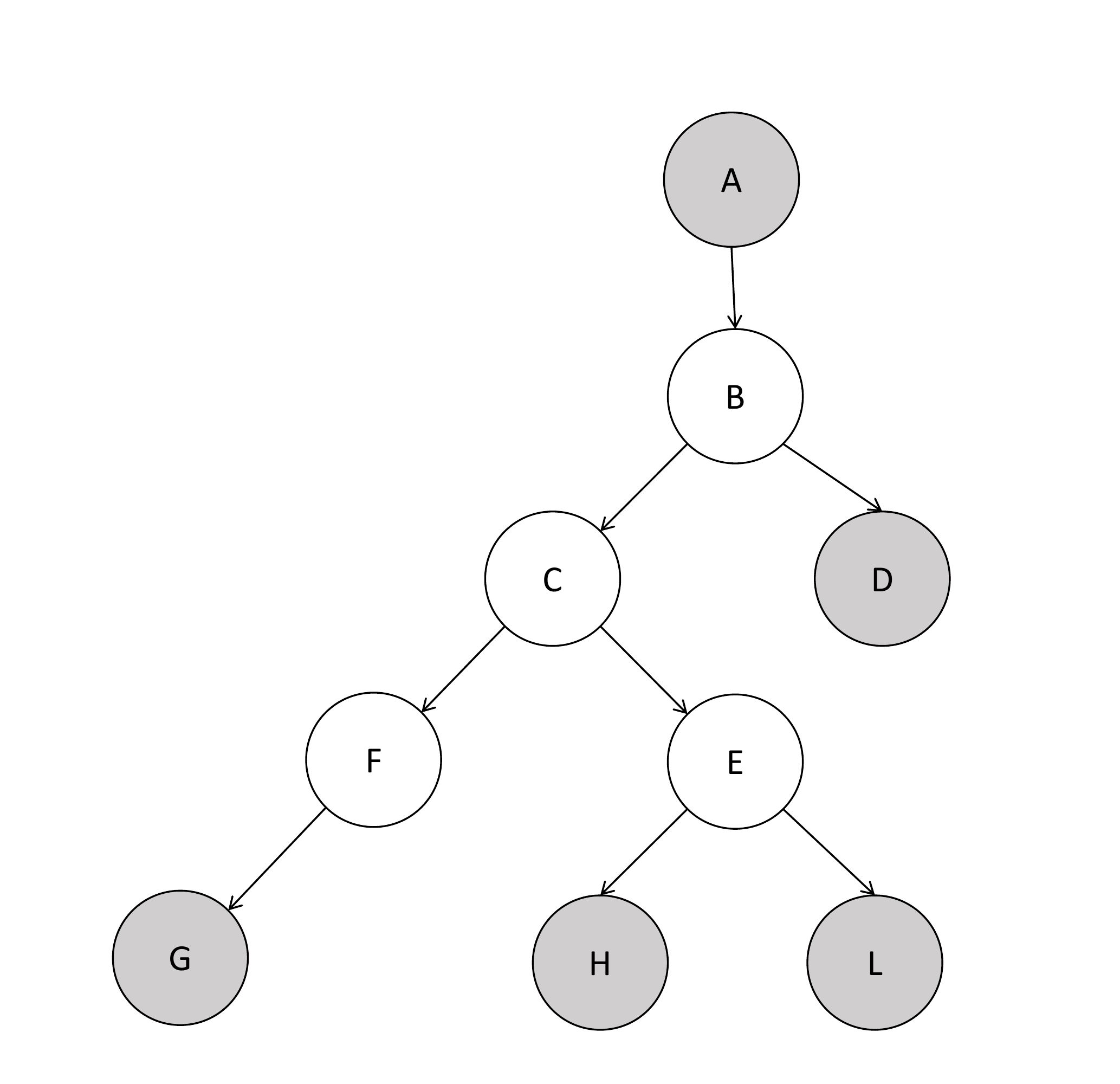}
        \caption{}
        \label{fig:net1}
    \end{subfigure}
    ~
    \begin{subfigure}[b]{0.25\textwidth}
        \includegraphics[width=\textwidth]{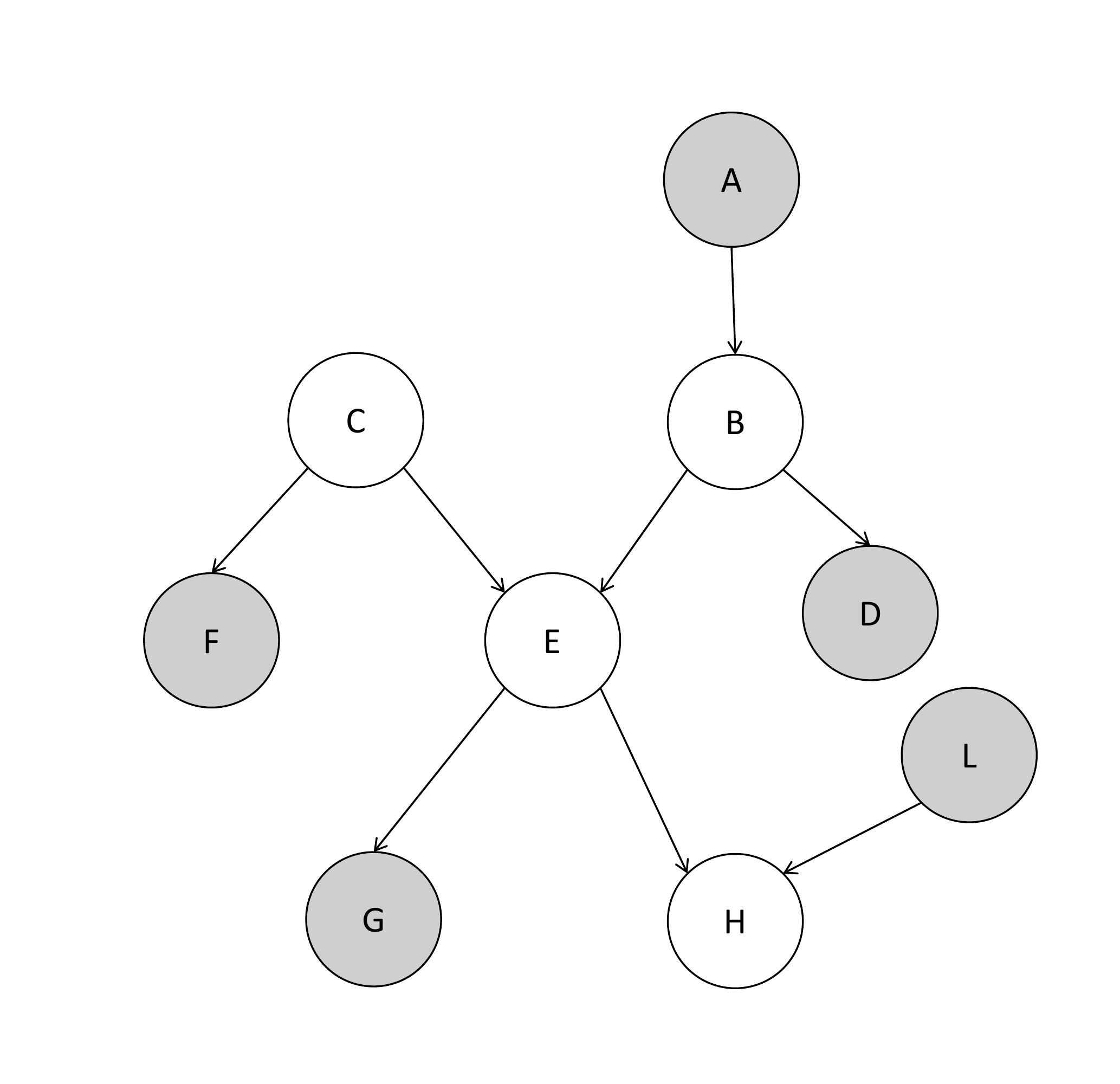}
        \caption{}
        \label{fig:net2}
    \end{subfigure}
    ~
    \begin{subfigure}[b]{0.25\textwidth}
        \includegraphics[width=\textwidth]{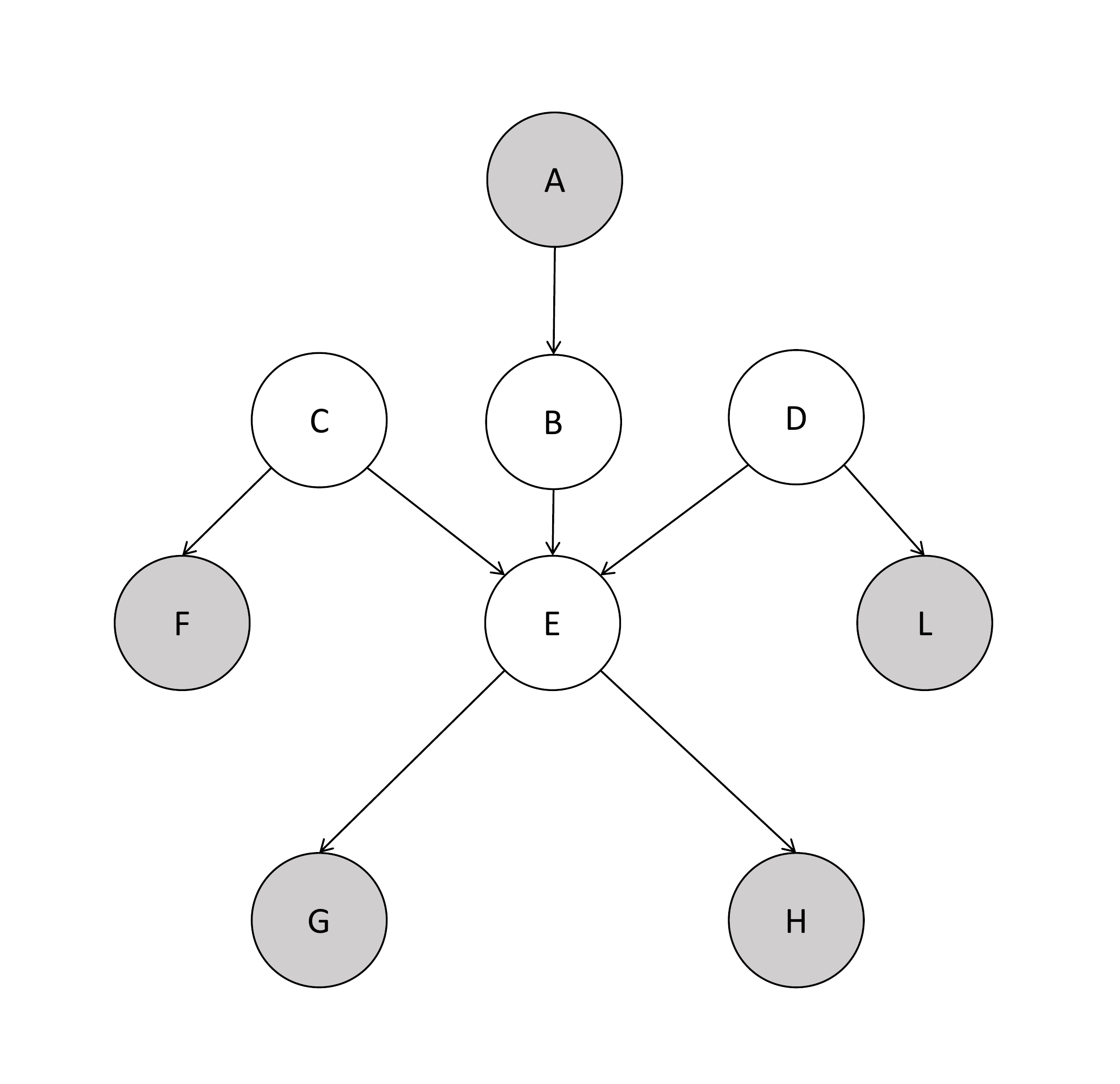}
        \caption{}
        \label{fig:net3}
    \end{subfigure}
    \caption{Network structures tested where the exterior gray variables are directly observed and the remaining are queried: (\subref{fig:net1}) Net1, a tree; (\subref{fig:net2}) Net2, singly connected network with one node having two parents; (\subref{fig:net3}) Net3, singly connected network with one node having three parents.}
    \label{fig:nets}
\end{figure*}

\begin{figure*}[t]
\centering
    \begin{subfigure}[b]{0.3\textwidth}
        \includegraphics[width=\textwidth]{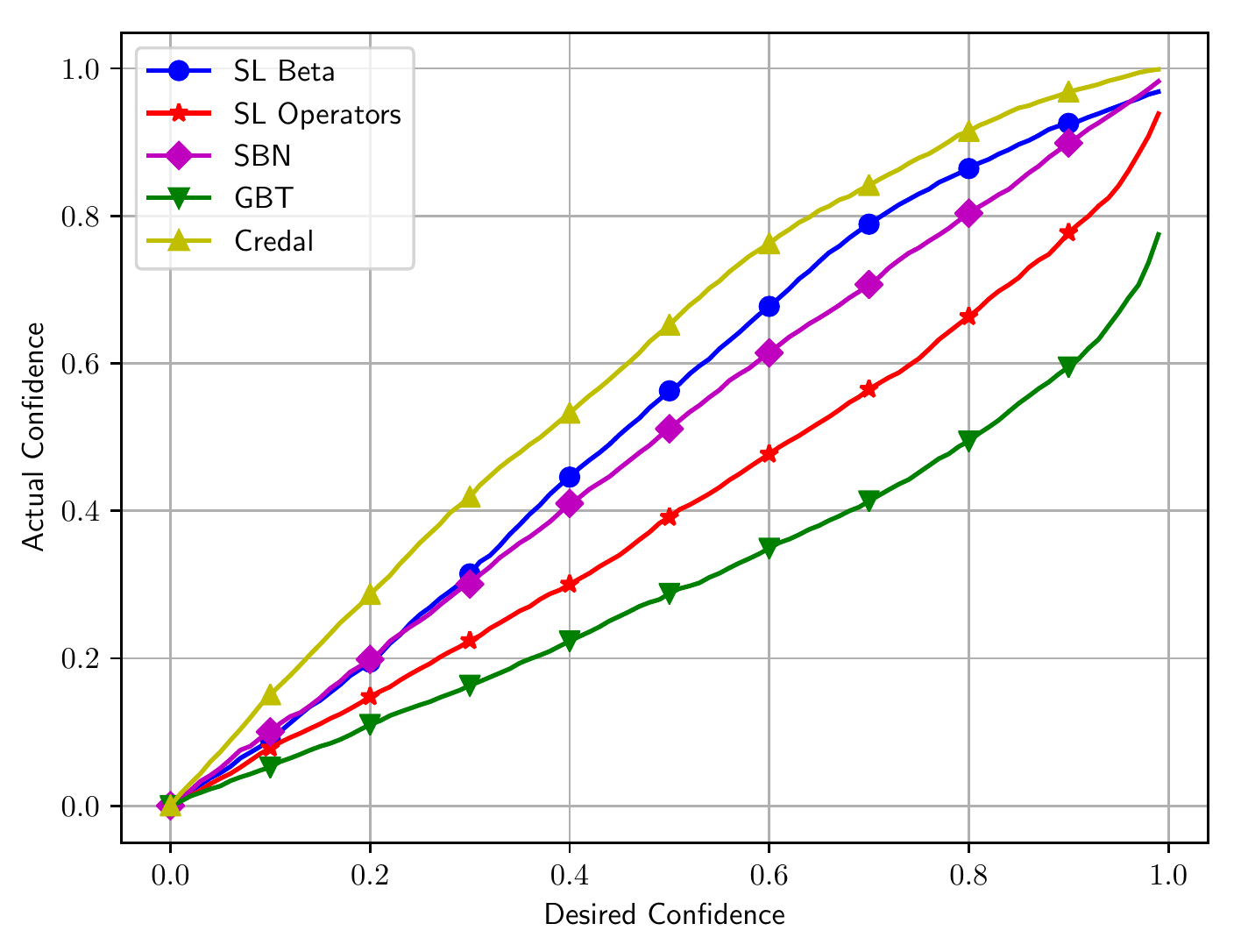}
        \caption{}
        \label{fig:net1a}
    \end{subfigure}
    ~
    \begin{subfigure}[b]{0.3\textwidth}
        \includegraphics[width=\textwidth]{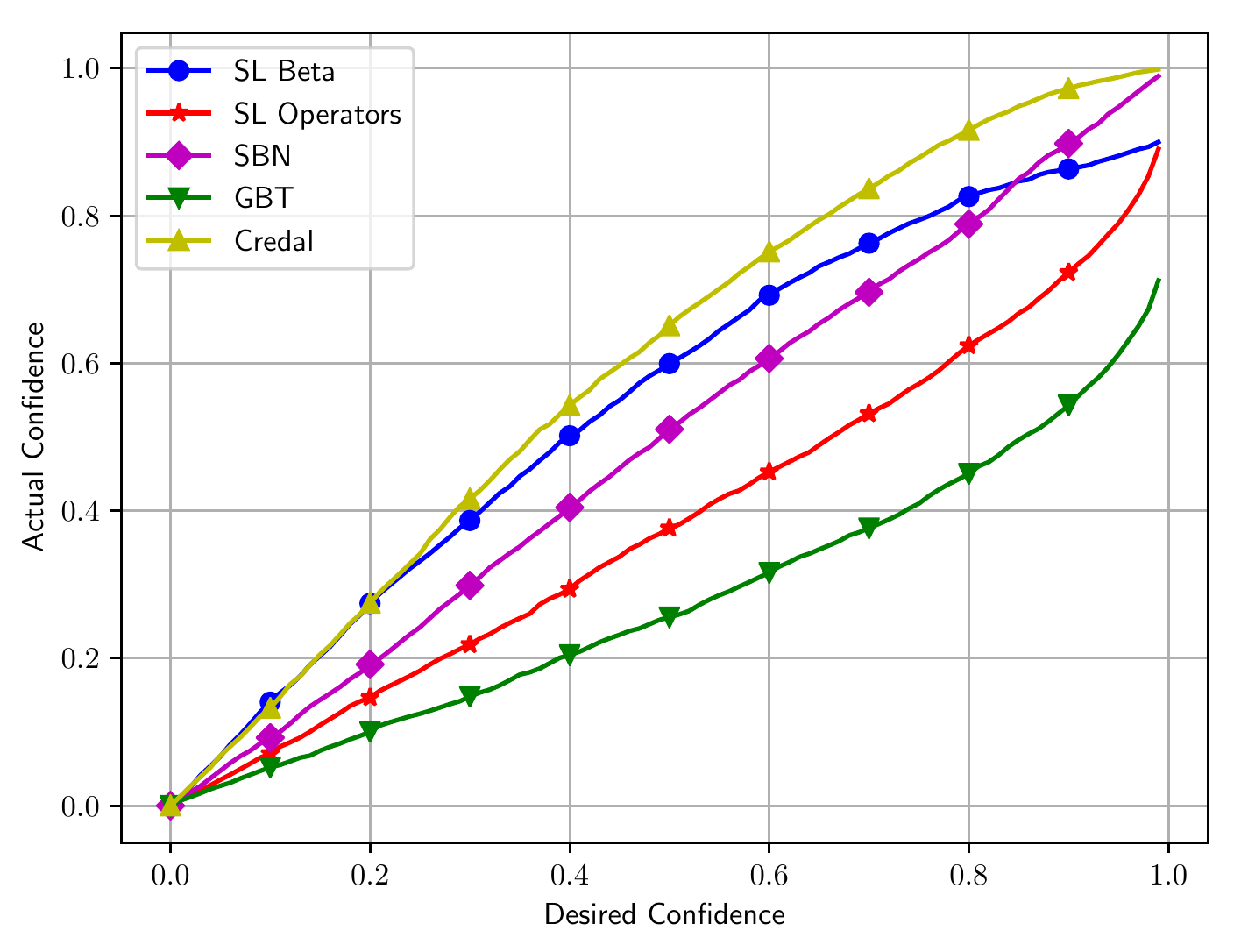}
        \caption{}
        \label{fig:net1b}
    \end{subfigure}
    ~
    \begin{subfigure}[b]{0.3\textwidth}
        \includegraphics[width=\textwidth]{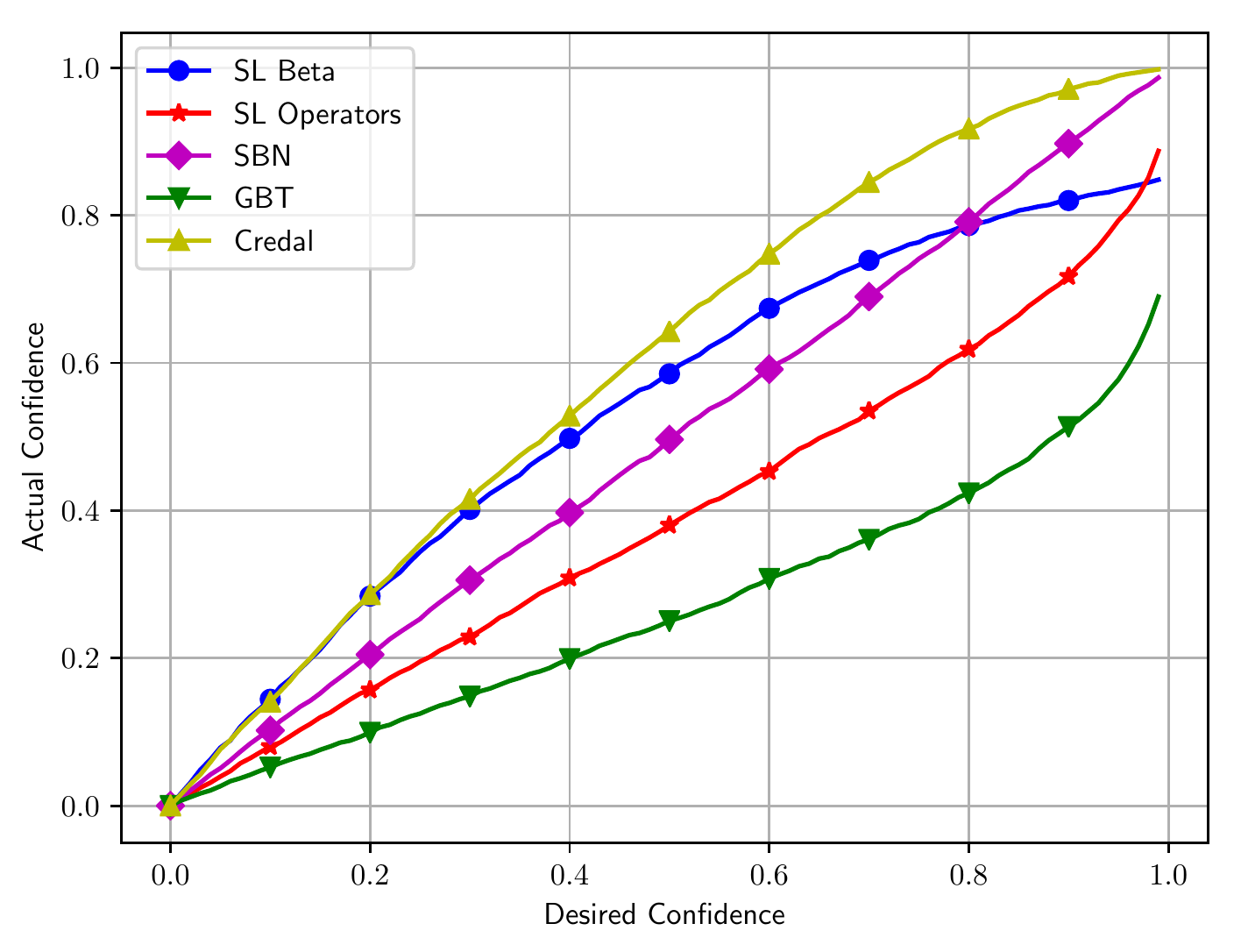}
        \caption{}
        \label{fig:net1c}
    \end{subfigure}
    
    \begin{subfigure}[b]{0.3\textwidth}
        \includegraphics[width=\textwidth]{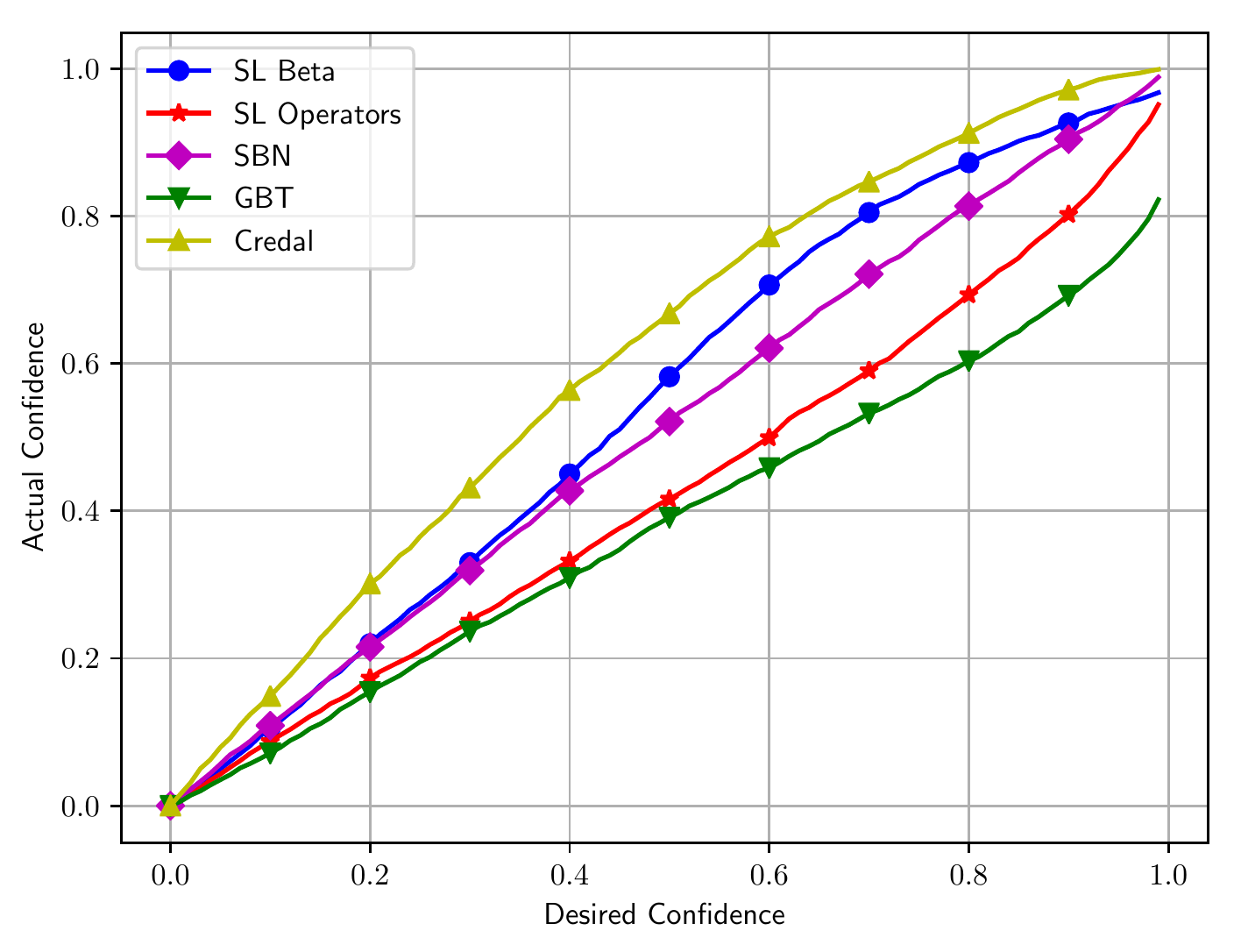}
        \caption{}
        \label{fig:net2a}
    \end{subfigure}
    ~
    \begin{subfigure}[b]{0.3\textwidth}
        \includegraphics[width=\textwidth]{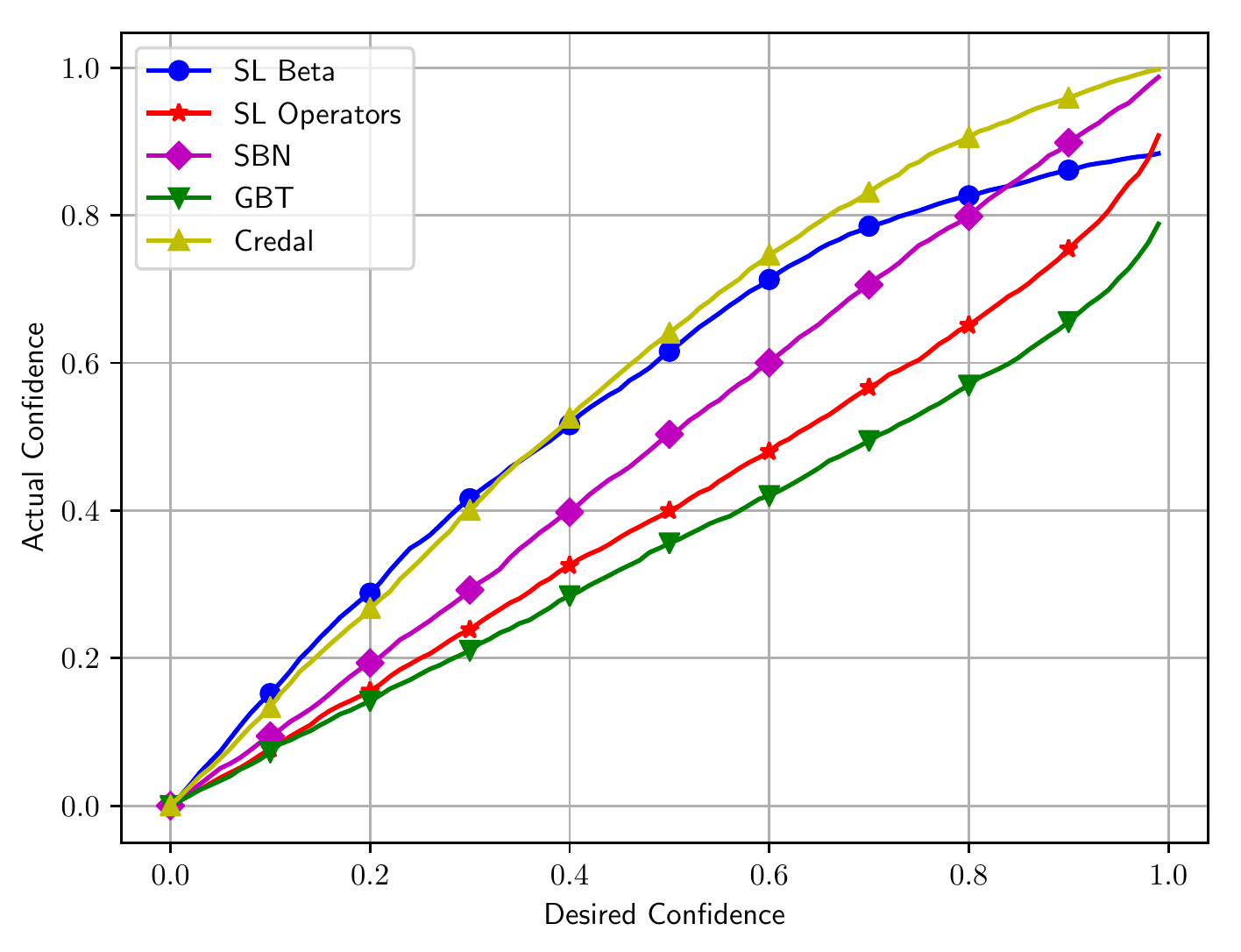}
        \caption{}
        \label{fig:net2b}
    \end{subfigure}
    ~
    \begin{subfigure}[b]{0.3\textwidth}
        \includegraphics[width=\textwidth]{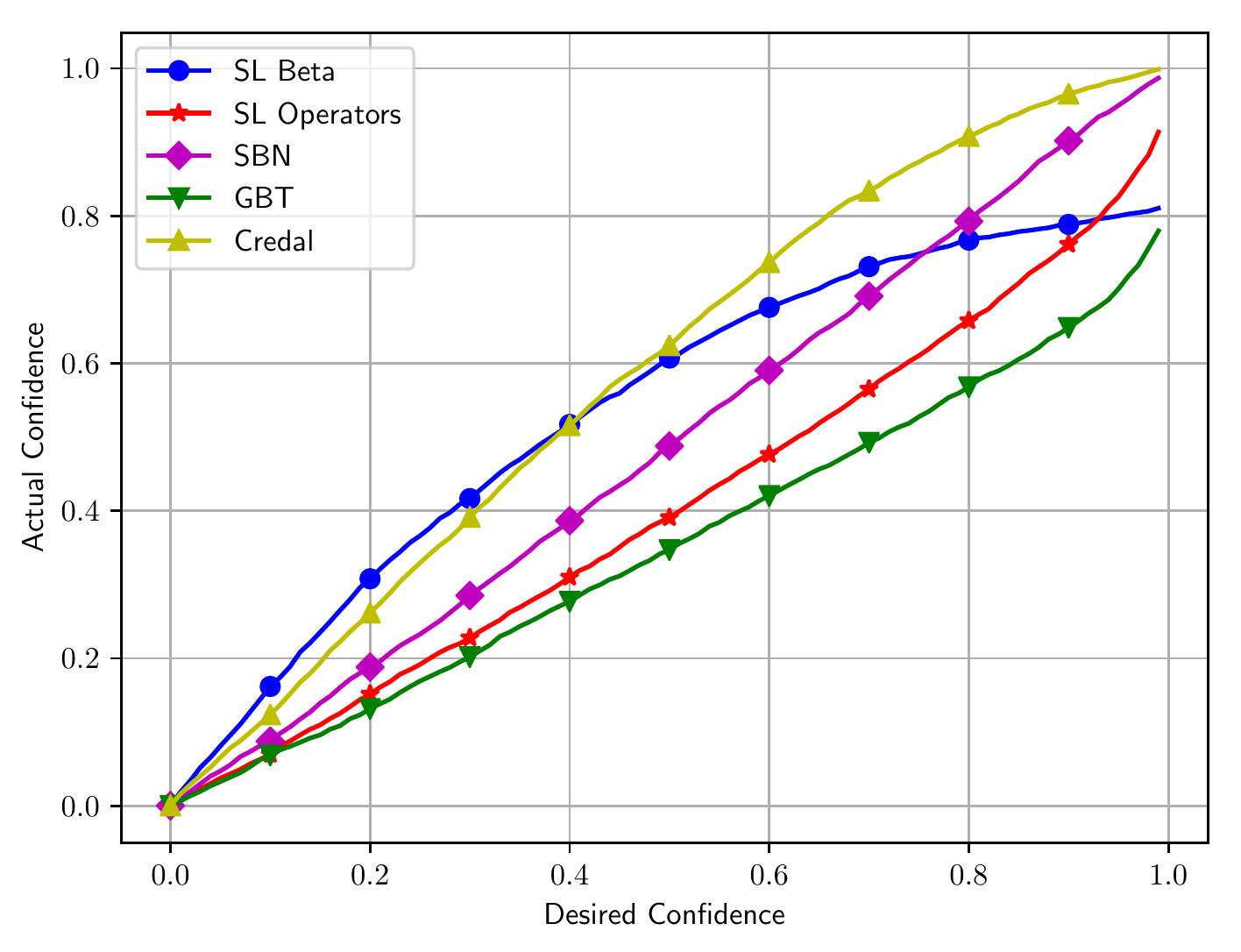}
        \caption{}
        \label{fig:net2c}
    \end{subfigure}
    
    \begin{subfigure}[b]{0.3\textwidth}
        \includegraphics[width=\textwidth]{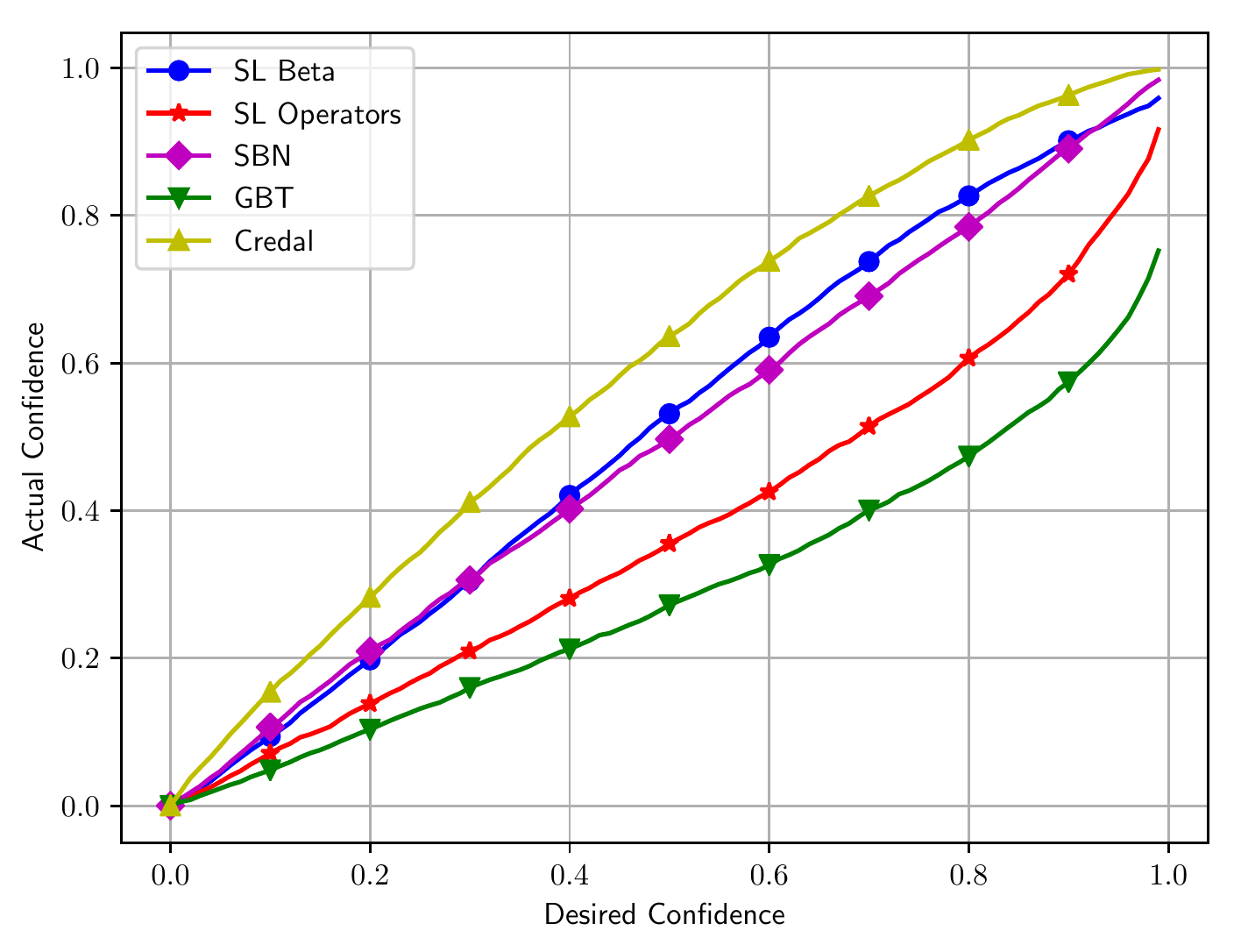}
        \caption{}
        \label{fig:net3a}
    \end{subfigure}
    ~
    \begin{subfigure}[b]{0.3\textwidth}
        \includegraphics[width=\textwidth]{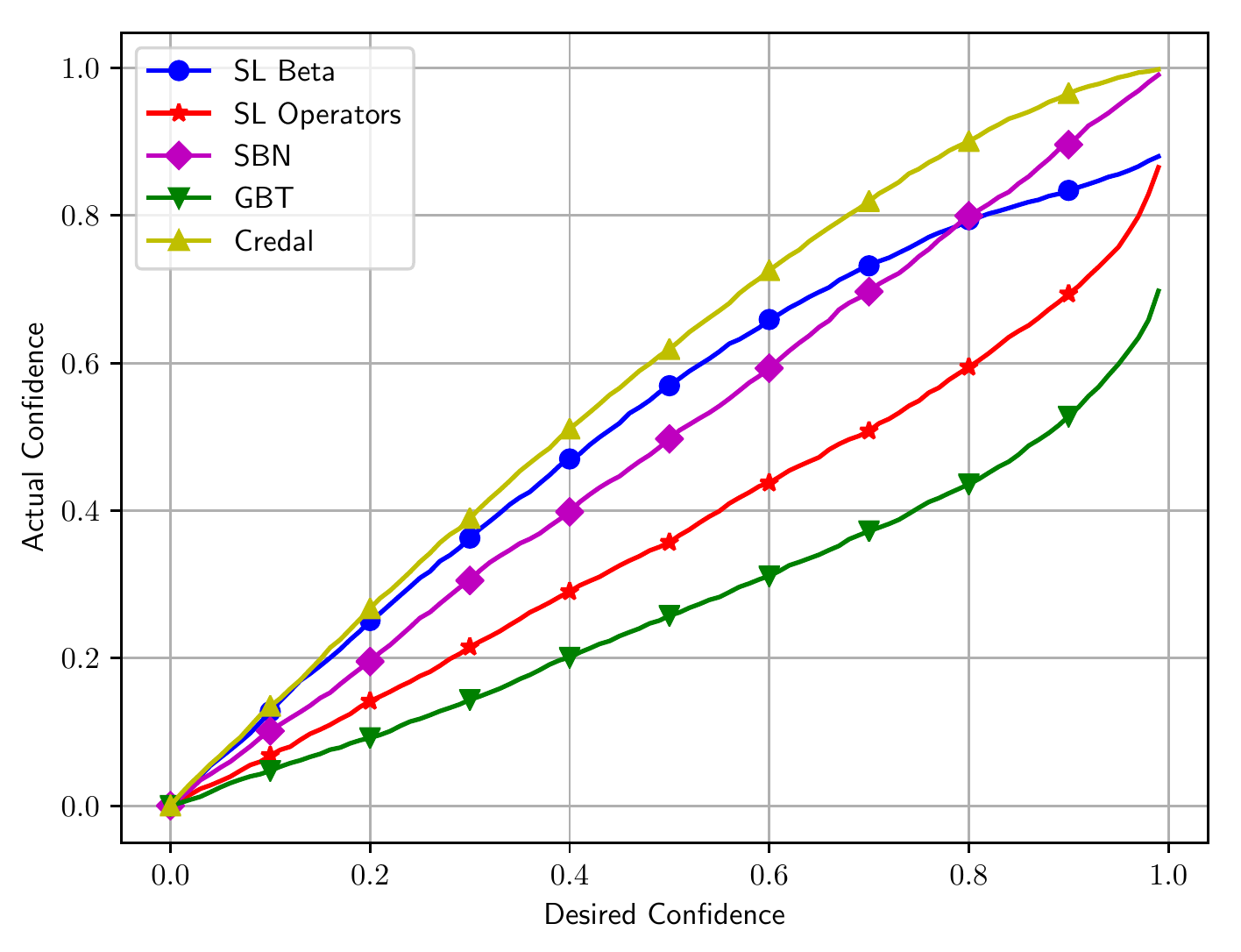}
        \caption{}
        \label{fig:net3b}
    \end{subfigure}
    ~
    \begin{subfigure}[b]{0.3\textwidth}
        \includegraphics[width=\textwidth]{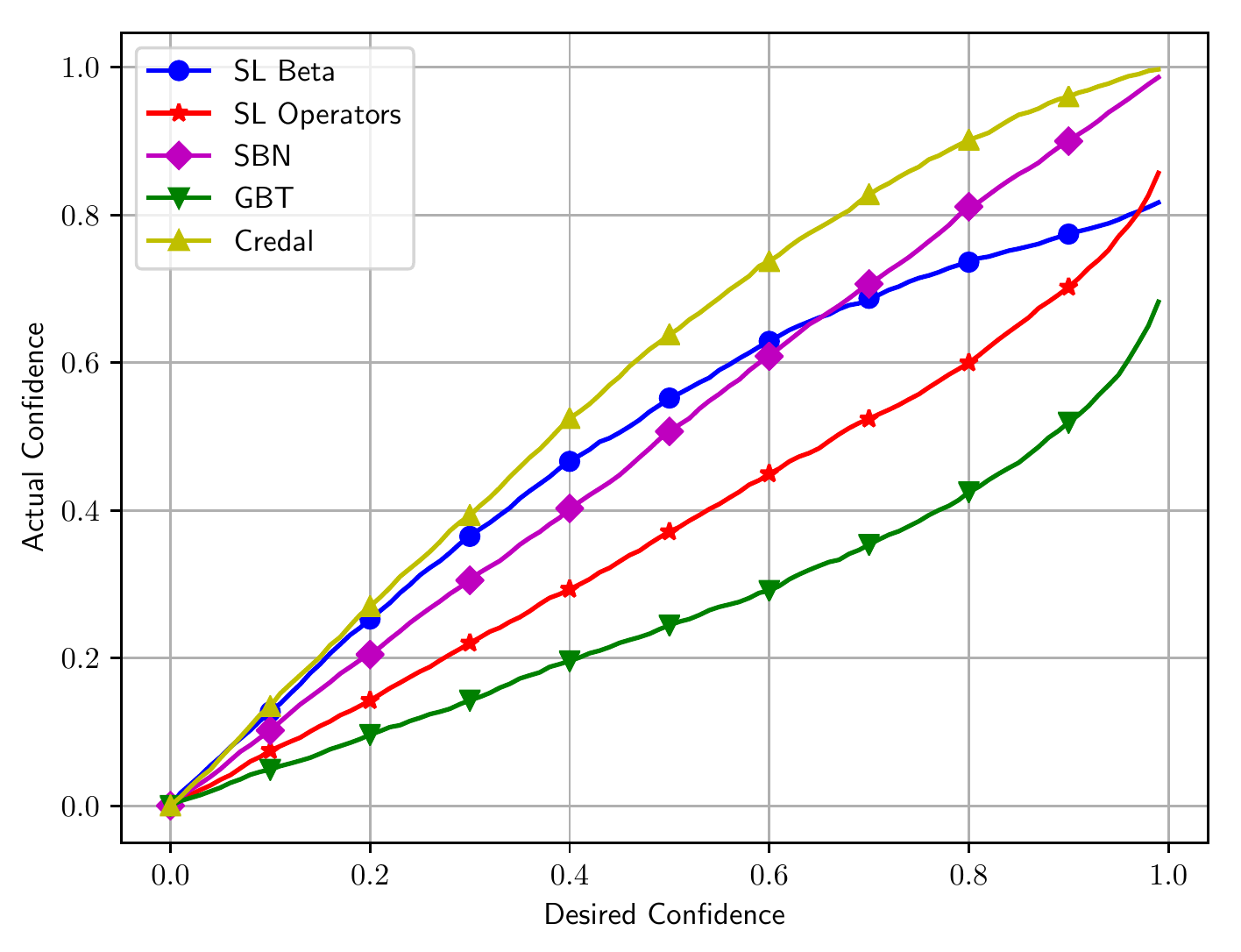}
        \caption{}
        \label{fig:net3c}
    \end{subfigure}
    
    \caption{Actual versus desired signiﬁcance of bounds derived from the uncertainty for: (\subref{fig:net1a}) Net1 with $N_{ins} = 10$; (\subref{fig:net1b}) Net1 with $N_{ins} = 50$; (\subref{fig:net1c}) Net1 with $N_{ins} = 100$; 
    (\subref{fig:net2a}) Net2 with $N_{ins} = 10$; (\subref{fig:net2b}) Net2 with $N_{ins} = 50$; (\subref{fig:net2c}) Net2 with $N_{ins} = 100$;
    (\subref{fig:net3a}) Net3 with $N_{ins} = 10$; (\subref{fig:net3b}) Net3 with $N_{ins} = 50$; (\subref{fig:net3c}) Net3 with $N_{ins} = 100$. Best closest to the diagonal. In the figure, \emph{SL Beta} represents aProbLog with $\semiringbeta$, and \emph{SL Operators} represents aProbLog with $\semiringsl$.}
    \label{fig:netsres}
\end{figure*}

We compared our approach against the state-of-the-art approaches for reasoning with uncertain probabilites---Subjective Bayesian Network \cite{ivanovska.15,kaplan.16.fusion,KAPLAN2018132}, Credal Network \cite{credal98}, and Belief Network \cite{Smets2005}---in the case that is handled by all of them, namely single connected Bayesian networks. We considered three networks proposed in \cite{KAPLAN2018132} 
that are depicted in Figure \ref{fig:nets}: from each network, we straightforwardly derived a aProbLog program.

\begin{table}
\small
\begin{tabu}{X[2,l] X[2,l] X[1,l] X[3,r] X[3,r] X[3,r] X[3,r] X[3,r]}
\toprule
 & $N_{ins}$ &   & $\semiringbeta$ & $\semiringsl$ & SBN    & GBT    & Credal\\
\midrule
Net1& 10& A & \textbf{0.1505} & 0.2078 & \textbf{0.1505} & 0.1530 & 0.1631 \\
& & P & 0.1994 & 0.1562 & 0.1470 & 0.0868 & 0.2009 \\
& 50& A & {0.0555} & 0.0895 & {0.0555} & 0.0619 & \textbf{0.0553} \\
& & P & 0.0950 & 0.0579 & 0.0563 & 0.0261 & 0.0761 \\
& 100& A & \textbf{0.0766} & 0.1182 & \textbf{0.0766} & 0.0795 & {0.0771} \\
& & P & 0.1280 & 0.0772 & 0.0763 & 0.0373 & 0.1028 \\
\dashedline
Net2& 10 & A & \textbf{0.1387} & 0.2089 & \textbf{0.1387} & 0.1416 & 0.1459 \\
& & P & 0.2031 & 0.1662 & 0.1391 & 0.1050 & 0.1849 \\
& 50& A & {0.0537} & 0.0974 & {0.0537} & 0.0561 & \textbf{0.0528} \\
& & P & 0.1002 & 0.0671 & 0.0520 & 0.0342 & 0.0683 \\
& 100& A & 0.0730 & 0.1229 & \textbf{0.0726} & 0.0752 & 0.0728 \\
& & P & 0.1380 & 0.0863 & 0.0725 & 0.0482 & 0.0949 \\
\dashedline
Net3& 10& A & 0.1566 & 0.2111 & \textbf{0.1534} & 0.1554 & 0.1643 \\
& & P & 0.1935 & 0.1517 & 0.1467 & 0.0832 & 0.1964 \\
& 50& A & 0.0697 & 0.0947 & \textbf{0.0548} & 0.0584 & \textbf{0.0548} \\
& & P & 0.0926 & 0.0602 & 0.0553 & 0.0242 & 0.0720 \\
& 100& A & 0.0879 & 0.1242 & 0.0745 & 0.0776 & \textbf{0.0743} \\
& & P & 0.1232 & 0.0798 & 0.0743 & 0.0347 & 0.0973 \\
\bottomrule
\end{tabu}
\caption{RMSE for the queried variables in the various networks: A stands for Actual, P for Predicted. Best results for the Actual RMSE in bold.}
    \label{tab:netsres}
\end{table}

As before, 
Table \ref{tab:netsres} provides the root mean square error (RMSE) between the projected probabilities and the ground truth probabilities for all the inferred query variables for $N_{ins}$ = 10, 50, 100, together with the RMSE predicted by taking the square root of the average variances from the inferred marginal Beta distributions. Figure \ref{fig:netsres} plots the desired and actual significance levels for the confidence intervals (best closest to the diagonal). 

Table \ref{tab:netsres} shows that aProbLog with $\semiringbeta$ shares the best performance with the state-of-the-art Subjective Bayesian Networks---in terms of actual RMSE---for Net1, and in two out of three cases of Net2 (all of them from a practical standpoint). This is clearly a significant achievement considering that Subjective Bayesian network is the state-of-the-art approach when dealing only with single connected Bayesian Networks with uncertain probabilities, while aProbLog with \semiringbeta\ can also handle much more complex problems. Net3 results are slightly worse due to approximations induced in the floating point operations used in the implementation: the more the connections of a node in the Bayesian network (e.g. node E in Figure \ref{fig:net3}), the higher the number of operations involved in \eqref{eq:fusion}. A more accurate code engineering can address it.
Consistently with Table \ref{tab:smokers}, aProbLog with $\semiringbeta$ has lower RMSE than with $\semiringsl$ and it underestimates its predicted RMSE, while aProbLog with $\semiringsl$ overestimates it. 

From visual inspection of Figure \ref{fig:netsres}, it is evident that aProbLog with $\semiringbeta$ performs best in presence of high uncertainty ($N_{ins} = 10$).
In presence of lower uncertainty, instead, it underestimates its own prediction up to a desired confidence between 0.6 and 0.8, and overestimate it after. This is due to the fact that aProbLog computes the conditional distributions at the very end of the process and \semiringbeta\ relies, in \eqref{eq:variance-division}, on the assumption that $X$ and $Y$ are uncorrelated.  However, since the correlation between $X$ and $Y$ is inversely proportional to $\sqrt{\var{X} \var{Y}}$, the lower the uncertainty, the less accurate our approximation.

\section{Conclusion}
We enabled the aProbLog approach to probabilistic logic programming to reason in presence of uncertain probabilities represented as Beta-distributed random variables. Other extensions to logic programming can handle uncertain probabilities by considering intervals of possible probabilities \cite{NG1992150}, similarly to the Credal network approach we compared against in Section \ref{sec:experiment}; or by
sampling random distributions, including ProbLog itself and cplint  \cite{alberti2017cplint} among others. Our approach does not require sampling or Monte Carlo computation, thus being significantly more efficient.

Our  experimental section shows that the proposed operators outperform the standard subjective logic operators and they are as good as the state-of-the-art approaches for uncertain probabilities in Bayesian networks while being able to handle much more complex problems.
Moreover, in presence of high uncertainty, which is our main research focus, the approximations we introduce in this paper are minimal, as Figures \ref{fig:net1a}, \ref{fig:net2a}, and \ref{fig:net3a} show, with the results of aProbLog with \semiringbeta\ being very close to the diagonal.

As part of future work we will (1) provide a different characterisation of the variance in \eqref{eq:variance-division} taking into consideration the correlation between $X$ and $Y$; (2) test the boundaries of our approximations to provide practitioners with pragmatic assessments and assurances; and (3) introduce an expectation-maximisation (EM) algorithm for learning labels representing Beta-distributed random variables with partial interpretations and compare it against the LFI algorithm \cite{gutmann2011learning} for ProbLog.

\appendix
\section{Subjective Logic Operators of Sum, Multiplication, and Division}
\label{sec:sl-operators}
  Let us recall the following operators as defined in \cite{Josang2016-SL}.
Let $\slop{X} = \sloptuple{X}$ and $\slop{Y} = \sloptuple{Y}$ be two subjective logic opinions, then:

\begin{itemize}
    \item the opinion about $X \cup Y$ (\textbf{sum}, $\slop{X} \boxplus_{\mbox{SL}} \slop{Y}$) is defined as $\slop{X \cup Y} = \sloptuple{X \cup Y}$, where
    $\slbel{X \cup Y} = \slbel{X} + \slbel{Y}$,
             $\sldis{X \cup Y} = \frac{\slbase{X} (\sldis{X}-\slbel{Y}) + \slbase{Y} (\sldis{Y} - \slbel{X})}{\slbase{X} + \slbase{Y}}$, 
             $\slunc{X \cup Y} = \frac{\slbase{X} \slunc{X} + \slbase{Y} \slunc{Y}}{\slbase{X} + \slbase{Y}}$, and
             $\slbase{X \cup Y} = \slbase{X} + \slbase{Y}$;

    \item the opinion about $X \land Y$ (\textbf{product}, $\slop{X} \boxtimes_{\mbox{SL}} \slop{Y}$) is defined---under assumption of independence---as $\slop{X \land Y} = \sloptuple{X\land Y}$, where
         $\slbel{X \land Y} = \slbel{X} \slbel{Y} + \frac{(1 - \slbase{X})\slbase{Y} \slbel{X} \slunc{Y} +\slbase{X}(1 - \slbase{Y})\slunc{X}\slbel{Y}}{1 - \slbase{X}\slbase{Y}}$,
     $\sldis{X \land Y} = \sldis{X}+{\sldis{Y}} - \sldis{X}\sldis{Y}$,
     $\slunc{X \land Y} = \slunc{X}\slunc{Y}+\frac{(1 - \slbase{Y})\slbel{X}\slunc{Y}+(1 - \slbase{X})\slunc{X}\slbel{Y}}{1 - \slbase{X}\slbase{Y}}$, and
     $\slbase{X \land Y} = \slbase{X} \slbase{Y}$;

    \item the opinion about the division of $X$ by $Y$, $X \widetilde{\land} Y$ (\textbf{division}, $\slop{X} \boxslash_{\mbox{SL}} \slop{Y}$) is defined as $\slop{X \widetilde{\land} Y} = \sloptuple{X \widetilde{\land} Y}$
    \slbel{X \widetilde{\land} Y} = 
             $\frac{\slbase{Y}(\slbel{X}+\slbase{X}\slunc{X})}{(\slbase{Y}-\slbase{X})(\slbel{Y}+\slbase{Y}\slunc{Y})}
             -
             \frac{\slbase{X}(1-\sldis{X})}{(\slbase{Y}-\slbase{X})(1-\sldis{Y})}
             $,
             $\sldis{X \widetilde{\land} Y} = \frac{\sldis{X}-\sldis{Y}}{1-\sldis{Y}}$, 
             $\slunc{X \widetilde{\land} Y} = 
             \frac{\slbase{Y}(1-\sldis{X})}{(\slbase{Y}-\slbase{X})(1-\sldis{Y})}
             -
             \frac{\slbase{Y}(\slbel{X}+\slbase{X}\slunc{X})}{(\slbase{Y}-\slbase{X})(\slbel{Y}+\slbase{Y}\slunc{Y})}
             $, and
             $\slbase{X \widetilde{\land} Y} = \frac{\slbase{X}}{\slbase{Y}}$, 

        subject to: $\slbase{X} < \slbase{Y}$; $\sldis{X} \geq \sldis{Y}$; 
        $\slbel{X} \geq \frac{\slbase{X}(1-\slbase{Y})(1-\sldis{X})\slbel{Y}}{(1-\slbase{X})\slbase{Y}(1-\sldis{Y})}$; 
        $\slunc{X} \geq \frac{(1-\slbase{Y})(1-\sldis{X})\slunc{Y}}{(1-\slbase{X})(1-\sldis{Y})}$.
\end{itemize}

\bibliography{biblio}
\bibliographystyle{apalike}
\end{document}